\documentclass[10pt,twocolumn,letterpaper]{article}

\usepackage{cvpr}
\usepackage{times}
\usepackage{epsfig}
\usepackage{graphicx}
\usepackage{amsmath}
\usepackage{amssymb}

\usepackage{times}
\usepackage{epsfig}

\usepackage{amsmath,bbm}
\usepackage{amssymb}
\usepackage{epstopdf}

\usepackage{enumerate}
\usepackage{subfigure}
\usepackage{tikz}
\usepackage{pgfplots}
\usepackage{pifont}
\usepackage{authblk}
\newcommand{\cmark}{\ding{51}}%

\cvprfinalcopy 

\def\cvprPaperID{****} 

\ifcvprfinal\fi
\begin{document}

\title{AugFPN: Improving Multi-scale Feature Learning for Object Detection}
\author[1]{Chaoxu Guo}
\author[1]{Bin Fan}
\author[2]{Qian Zhang}
\author[1]{Shiming Xiang}
\author[1]{Chunhong Pan}
\affil[1]{NLPR,CASIA}
\affil[2]{Horizon Robotics}
\affil[1]{\{chaoxu.guo, bfan, smxiang, chpan\}@nlpr.ia.ac.cn}
\affil[2]{qian01.zhang@horizon.ai}

\maketitle

\begin{abstract}
Current state-of-the-art detectors typically exploit feature pyramid to detect objects at different scales. Among them, FPN is one of the representative works that build a feature pyramid by multi-scale features summation. However, the design defects behind prevent the multi-scale features from being fully exploited. In this paper, we begin by first analyzing the design defects of feature pyramid in FPN, and then introduce a new feature pyramid architecture named Augmented FPN (AugFPN) to address these problems. Specifically, AugFPN consists of three components: Consistent Supervision, Residual Feature Augmentation, and Soft RoI Selection. AugFPN narrows the semantic gaps between features of different scales before feature fusion through Consistent Supervision. In feature fusion, ratio-invariant context information is extracted by Residual Feature Augmentation to reduce the information loss of feature map at the highest pyramid level. Finally, Soft RoI Selection is employed to learn a better RoI feature adaptively after feature fusion. By replacing FPN with AugFPN in Faster R-CNN, our models achieve 2.3 and 1.6 points higher Average Precision (AP) when using ResNet50 and MobileNet-v2 as backbone respectively. Furthermore, AugFPN improves RetinaNet by 1.6 points AP and FCOS by 0.9 points AP when using ResNet50 as backbone. Codes will be made available.
\end{abstract}

\section{Introduction}
    With the significant advances in deep convolutional networks (ConvNets), remarkable progress has been achieved in image object detection. A number of detectors \cite{rcnn,faster-rcnn,fast-rcnn,ssd,yolo,mask-rcnn,fpn,retinanet} have been proposed to steadily push forward the state-of-the-art. Among these detectors, FPN \cite{fpn} is a simple and effective two-stage framework for object detection. Specifically, FPN builds a feature pyramid upon the inherent feature hierarchy in ConvNet by propagating the semantically strong features from high levels into features at lower levels.

   By improving multi-scale features with strong semantics, the performance of object detection has been substantially improved. However, there exist some design defects within the feature pyramid in FPN, which is illustrated in Fig. \ref{fig0}. Basically, the feature pyramid in FPN can be formulated into three stages: (1) before feature fusion, (2) top-down feature fusion, and (3) after feature fusion. We find that each stage has an intrinsic flaw as described in the following:

\begin{figure}
    \includegraphics[width=1.1\columnwidth]{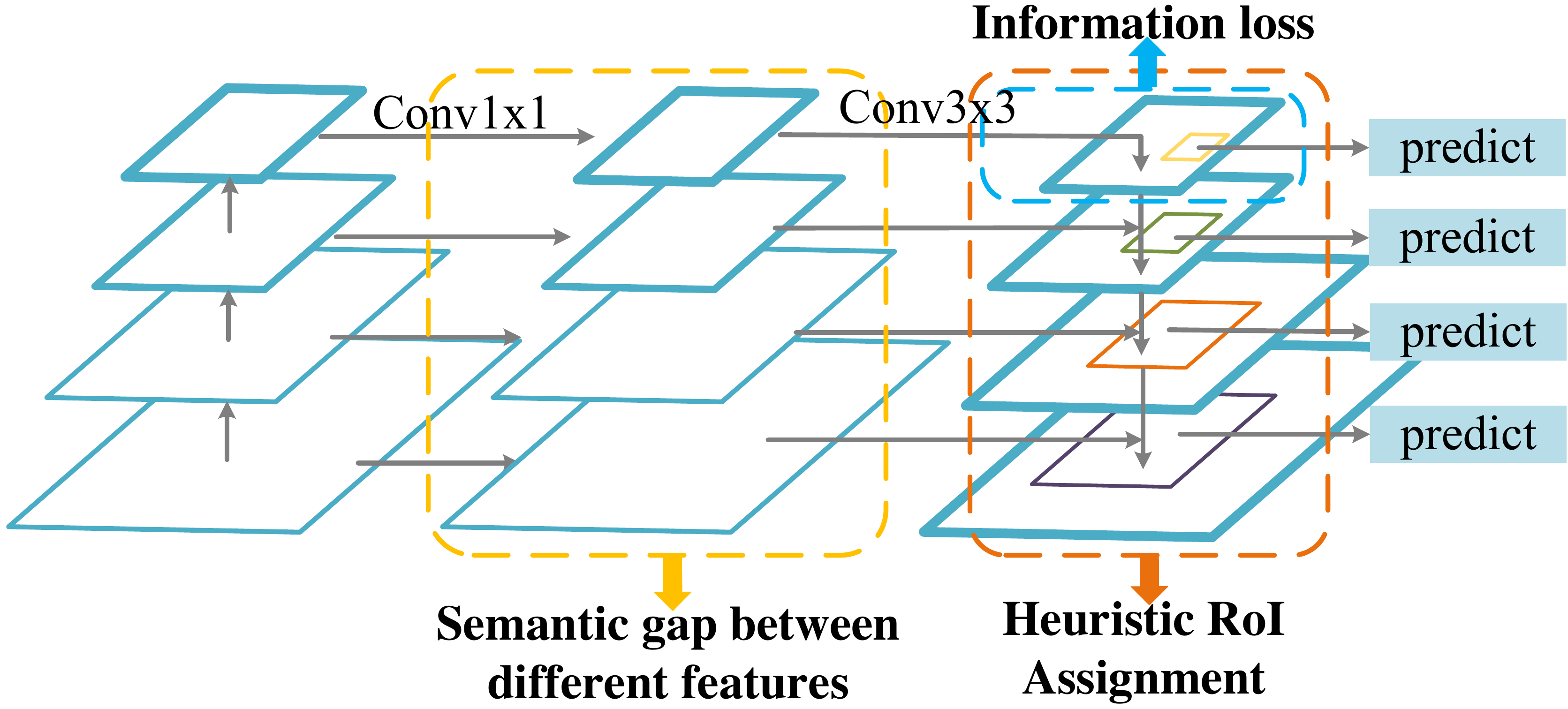}
    \label{fig0}
    \caption{Three design defects in feature pyramid network: 1) \textbf{semantic gap between features at different levels} before feature summation, 2)\textbf{ information loss} of the feature at the highest pyramid level, 3) \textbf{heuristic RoI assignment}.}
\end{figure}

   \paragraph{Semantic gaps between features at different levels.} Before performing feature fusion, features at different levels undergo a $1\times 1$ convolution layer independently to reduce feature channels, where the large semantic gaps between these features are not considered. Fusing these features directly would degrade the power of multi-scale feature representation due to the inconsistent semantic information.

   \paragraph{Information loss of the highest-level feature map.} In feature fusion, features are propagated in a top-down path and low-level features can be improved with the stronger semantic information from higher-level features. Nevertheless, the feature at the highest pyramid level instead loses information due to the reduced channels. The information loss can be mitigated by combining the global context feature \cite{thundernet} extracted by global pooling. But such a strategy of fusing the feature map into one single vector may lose the spatial relation and details because multiple objects may appear in one image.

   \paragraph{Heuristical assignment strategy of RoIs .} After feature fusion, each object proposal is refined based on the feature grids pooled from one feature level, which is chosen based on the scales of proposals heuristically. However, the ignored features from other levels may be beneficial for object classification or regression. Considering this problem, PANet \cite{panet} pools RoIs features from all pyramid levels and fuses them with the max operation after adapting them with independent fully connected layers. Nevertheless, the max fusion would ignore features with smaller responses that may be also helpful and still does not exploit the features at other levels fully. Meanwhile, the extra fully connected layers increase the model parameters significantly.

   In this paper, we propose AugFPN, a simple yet effective feature pyramid that integrates three different components to deal with the problems above respectively. First, Consistent Supervision is proposed to make the feature maps after lateral connection contain similar semantic information by enforcing the same supervision signals on these feature maps. Second, ratio-invariant adaptive pooling is utilized to extract diverse context information, which could reduce information loss of the highest-level feature in feature pyramid in a residual way. We name this procedure as Residual Feature Augmentation. Third, Soft RoI Selection is introduced to better exploit RoI features from different pyramid levels and produce a better RoI feature for subsequent location refinement and classification.

   Without bells and whistles, AugFPN based Faster R-CNN outperforms FPN based counterparts by 2.3 and 1.7 Average Precision (AP) when using ResNet50 and ResNet101 as backbone respectively. Furthermore, AugFPN improves the overall performance by 1.6 AP when the backbone is changed to MobileNet-V2, which is a light-weight and efficient network. AugFPN can also be extended to one-stage detectors with minor modifications. By replacing FPN with AugFPN, RetinaNet and FCOS are improved by 1.6 AP and 0.9 AP respectively, which manifests the generality of AugFPN.

   We summarize our contributions as follows:
   \begin{itemize}
      \item  We reveal the issues in three different stages of FPN that prevent the multi-scale features from being fully exploited.
      \vspace{-0.2cm}
      \item  A new feature pyramid network named AugFPN is proposed to address these problems with Consistent Supervision, Residual Feature Augmentation, and Soft RoI Selection respectively.
      \vspace{-0.2cm}
      \item  We evaluate AugFPN equipped with various detectors and backbones on MS COCO and it consistently brings significant improvements over FPN based detectors.
   \end{itemize}

\section{Related Work}
   \paragraph{Deep Object Detectors.} Contemporary object detection methods almost follow two paradigms, two-stage and one-stage. As a seminal work of the two-stage detection methods \cite{rcnn,fast-rcnn,faster-rcnn,rfcn,fpn,cascade,snip,trident,lighthead,libra}, R-CNN \cite{rcnn} first employs selective search \cite{selective} to generate region proposals and then refines these proposals by extracting region features through a convolutional network. To improve the training and inference speed, SPP \cite{spp} and Fast R-CNN \cite{fast-rcnn} first extract feature map of the whole image and then generate region features with spatial pyramid pooling and RoI pooling respectively. Finally the region fetures are used to refine the proposals. Faster R-CNN \cite{faster-rcnn} proposes a region proposal network and develops an end-to-end trainable detector, which promotes the performance significantly and speed-up the inference. To pursue scale-invariance in object detection, FPN \cite{fpn} builds an in-network feature pyramid based on the inherent feature hierarchy of convolution network and makes predictions at different pyramid levels according to the scales of region proposals. RoI Align \cite{mask-rcnn} brings great improvement in both object detection and instance segmentation by addressing the quantization problem of RoI pooling. Deformable network \cite{deformable, deformablev2}  improve the performance of object detection significantly by modeling the geometry structure of objects. Cascade R-CNN \cite{cascade} introduces a multi-stage refinement into Faster R-CNN and achieves more accurate predictions of object locations.

   Contrary to two-stage detectors, one-stage detectors \cite{ssd,yolo,dssd,yolov2,retinanet,cornernet,rfbnet,yolov3,refinedet,extremenet} are more efficient yet less accurate. SSD \cite{ssd} places anchor boxes densely on multi-scale features and make predictions based on these anchors. RetinaNet \cite{retinanet} utilizes a feature pyramid similar to FPN as backbone and introduces a novel focal loss to address the imbalance problem of easy and hard examples. ExtremeNet\cite{extremenet} models the problem of object detection as detecting four extreme points of the objects. These works make significant progress from different concerns. In this paper, we study a better exploitation of multi-scale features.

\begin{figure*}
    \includegraphics[width=1\textwidth]{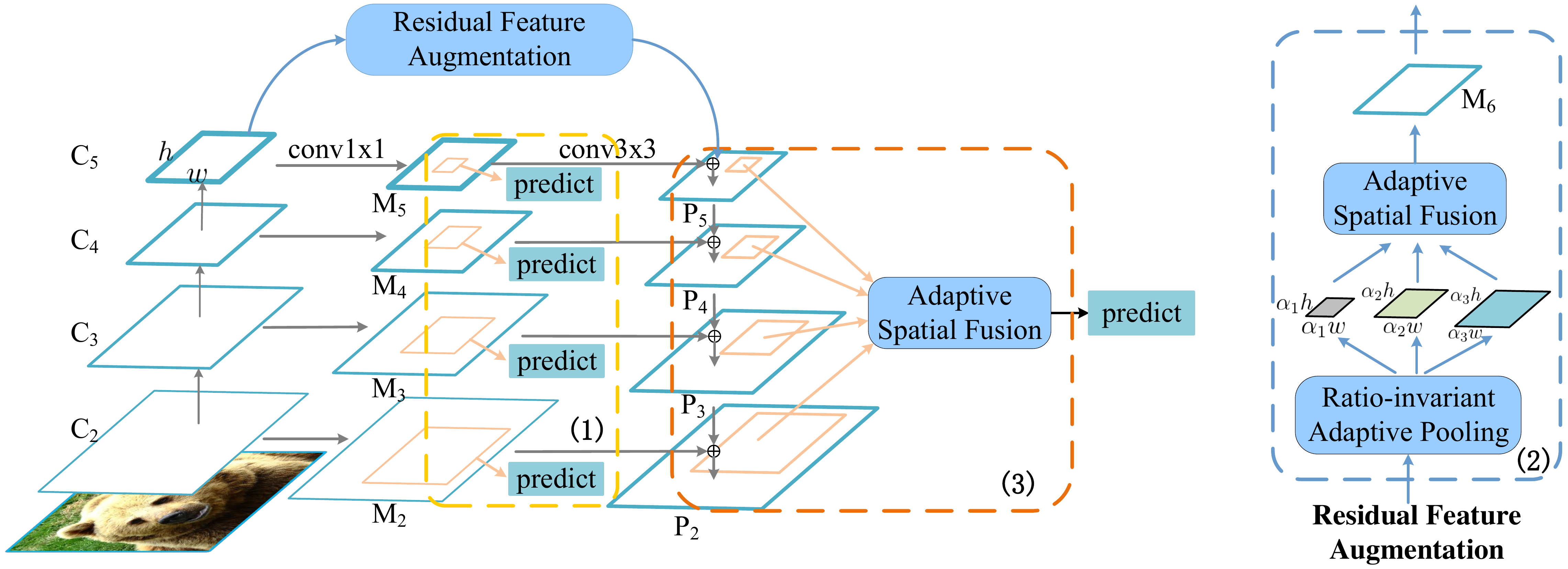}
    \label{fig1}
    \caption{Overall pipeline of AugFPN based detector. (1)-(3) are three main components of AugFPN: Consistent Supervision, Residual Feature Augmentation, and Soft RoI Selection. For simplicity, the $3\times3$ convolution layers after feature summation are not shown.}
\end{figure*}

   \paragraph{Deep Supervision.} Deep supervision \cite{multi,dsn,psp,nasfpn} is a wildly used technique to tackle the common problem of gradient vanishing or enhance the feature representation of intermediate layers. Huang \textit{et al.} \cite{multi} incorporate several classifiers with various resource demands into a single deep network by training it at different levels simultaneously. PSPNet \cite{psp} introduces an additional pixel-level loss on intermediate layers in order to reduce the optimizing difficulty. Recently Nas-FPN \cite{nasfpn} attaches classifier and regression heads after all intermediate pyramid networks with a goal of achieving \textit{anytime detection}. Contrary to these works, we apply the instance-level supervision signals on features at all pyramid levels after lateral connection, aiming to narrow the semantic gaps between them and make the features more suitable for subsequent feature summation.

   \paragraph{Context Exploitation.} Several methods have proved the importance of context on both object detection \cite{gidaris2015object,thundernet,gbd} and segmentation \cite{ccnet,parsenet,psp}. Deeplab-v2 \cite{deeplabv2} proposes atrous convolution to extract multi-scale context and PSPNet \cite{psp} utilizes pyramid pooling to obtain hierarchical global context, both of which improve the quality of semantic segmentation greatly. Different from them, we perform ratio-invariant adaptive pooling to generate diverse spatial context information and utilize them to reduce information loss in channels of the feature at the highest pyramid level in a residual way.

   \paragraph{Strategy of RoI Assignment.} FPN \cite{fpn} pools RoI features from one certain pyramid level, which is chosen according to the scales of RoIs. However, two proposals with a similar scale can be assigned to different feature levels under this strategy, which may produce sub-optimal results. To address this, PANet pools RoI features from all pyramid levels and fuses them by max operation after adapting them with a fully connected layer independently. There is a distinct difference between PANet and our work that we propose a data-dependent way to generate adaptive weights and absorb features from all levels according to the weights. This enable the features at different levels to be better exploited. In addition, our work requires fewer parameters because no extra fully connected layers are required to adapt RoI features.

\begin{figure*}
    \centering
    \subfigure[]{
    \includegraphics[width=1\columnwidth]{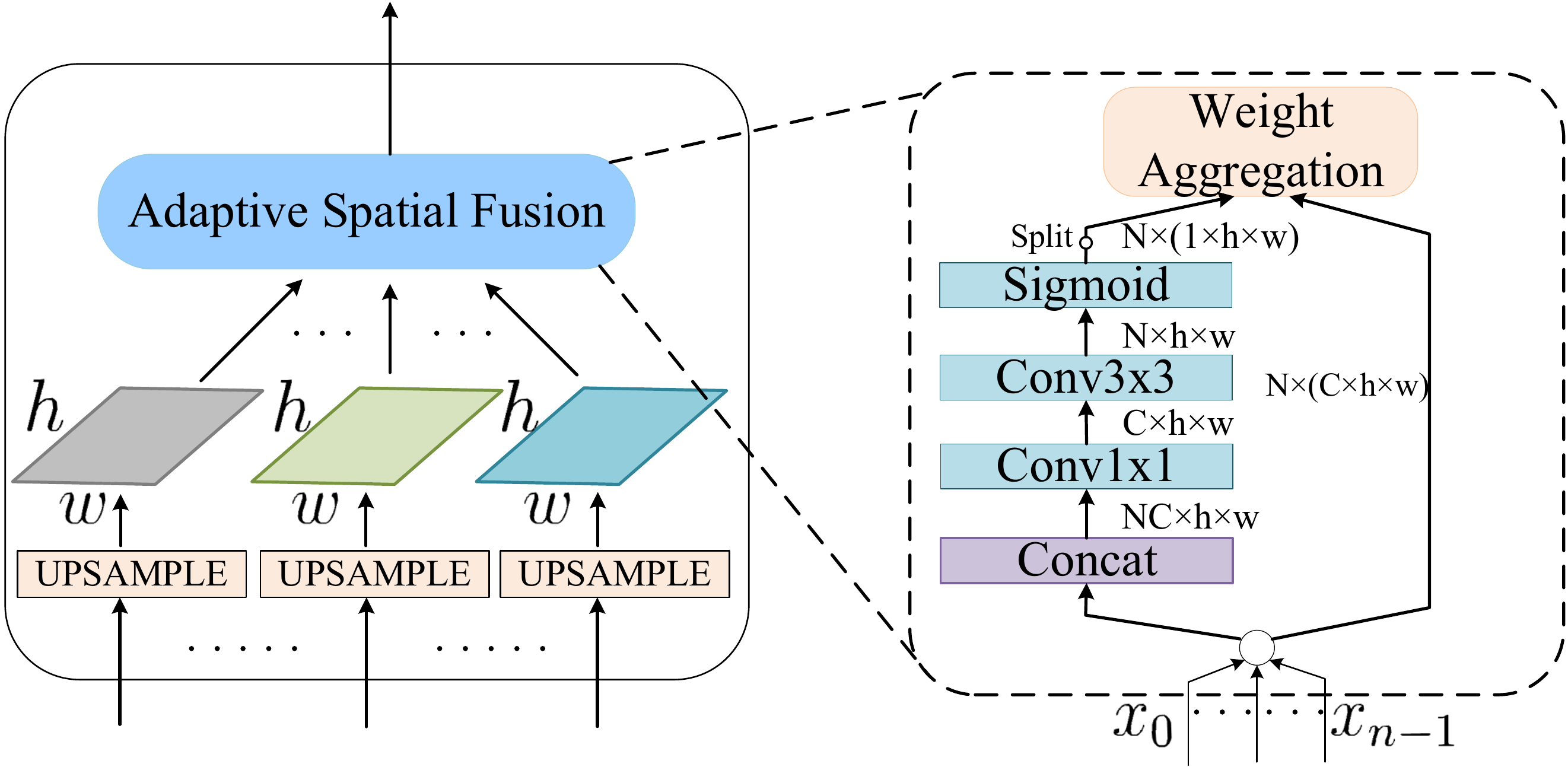}
    \label{fig2.1}}
    \subfigure[]{
    \includegraphics[width=1\columnwidth]{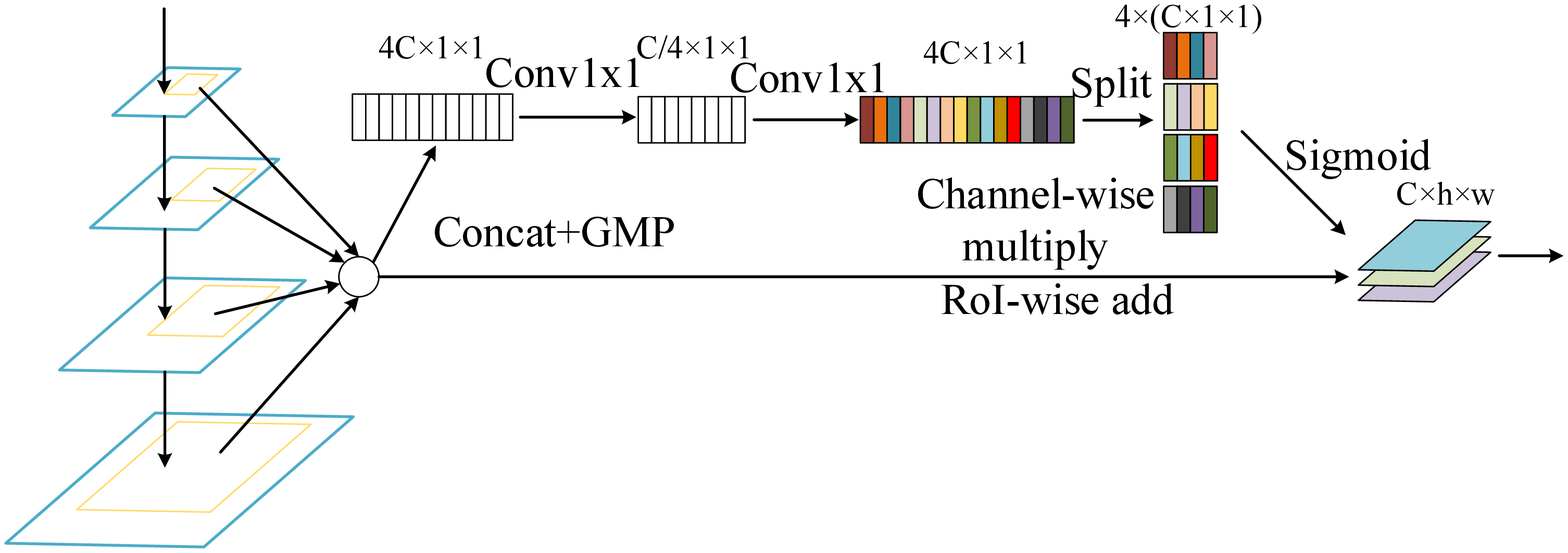}
    \label{fig2.2}}
    \caption{(a) is the process of fusing different context features and structure of Adaptive Spatial Fusion. (b) is the details of Adaptive Channel Fusion.}
\end{figure*}

\section{Methodology}
   The overall framework of AugFPN is shown in Figure \ref{fig1}. Following the setting of FPN \cite{fpn}, features used to build the feature pyramid are denoted as $\left\{C_2, C_3, C_4, C_5\right\}$, which correspond to the feature maps with strides $\left\{4, 8, 16, 32\right\}$ pixels in feature hierarchy w.r.t. the input image. $\left\{M_2, M_3, M_4, M_5\right\}$ are the features with reduced feature channels after lateral connection. $\left\{P_2, P_3, P_4, P_5\right\}$ are the features produced by feature pyramid. Three components of AugFPN will be discussed in the following subsections.

\subsection{Consistent Supervision}\label{cs}

   FPN makes use of the in-network feature hierarchy that produces feature maps with different resolutions to build a feature pyramid. In order to integrate the multi-scale context information, FPN fuses features of different scales by upsampling and summation in a top down path. However, the features with different scales contain information at different abstract levels and there exist large semantic gaps between them. Although the method adopted by FPN is simple and effective, fusing multiple features with large semantic gaps would lead to a sub-optimal feature pyramid.

   This inspires us to propose Consistent Supervision, which enforces the same supervision signals on the multi-scale features before fusion, with the goal of narrowing semantic gaps between them. Specifically, we first build a feature pyramid based on the multi-scale features $\left\{C_2, C_3, C_4, C_5\right\}$ from backbone. Then a Region Proposal Network (RPN) is appended to the resulting feature pyramid $\left\{P_2, P_3, P_4, P_5\right\}$ to generate numerous RoIs. To conduct Consistent Supervision, each RoI is mapped to all feature levels and the RoI features from each level of $\left\{M_2, M_3, M_4, M_5\right\}$ are obtained by RoI-Align \cite{mask-rcnn}. After that, multiple classification and box regression heads are attached to these features to generate auxiliary loss. The parameters of these classification and regression heads are shared across different levels, which can further force different feature maps to learn similar semantic information besides the same supervision signals. For more stable optimization, a weight is used to balance the auxiliary loss generated by Consistent Supervision and the original loss. Formally, the final loss function of rcnn head is formulated as follows:

   \begin{equation}
   \begin{aligned}
       L_{rcnn} = \lambda (L_{cls,M}(p_M,t^*) + \beta[t^*>0]L_{loc, M}(d_M, b^*))\\  + L_{cls,P}(p,t^*) + \beta[t^*>0]L_{loc, P}(d, b^*).
   \end{aligned}
   \label{equ1}
   \end{equation}
   $L_{cls,M}$ and $L_{loc,M}$ are objective functions corresponding to the auxiliary loss attached to $\left\{M_2, M_3, M_4, M_5\right\}$ while $L_{cls,P}$ and $L_{loc,P}$ are original loss functions on feature pyramid $\left\{P_2, P_3, P_4, P_5\right\}$. $p_M, d_M$ and $p, d$ are the prediction of intermediate layers and final pyramid layers respectively. $t^*$ and $b^*$ are the groundtruth class label and regression target respectively. $\lambda$ is the weight used to balance the auxiliary loss and original loss. $\beta$ is the weight used to balance classification and localization loss. The definition of $[t^*>0]$ is as follows:

   \begin{equation}
       [t^*>0] = \left\{
       \begin{aligned}
       1, t^* > 0  \\
       0, t^* =0 \\
       \end{aligned}
       \right.
   \end{equation}

   In the testing phase, the auxiliary branches are abandoned and only the branch after feature pyramid is utilized for final prediction. Therefore, Consistent Supervision introduces no extra parameters and computation to the model in inference.

   \begin{table*}
   \centering
   \resizebox{1.0\textwidth}{!}{
   \begin{tabular}{ccccccccc}
   \hline
   Method                  &      Backbone          &  Schedule        & AP     & $AP_{50}$    & $AP_{75}$     & $AP_{S}$      & $AP_{M}$       & $AP_{L}$     \\
   YoLOv2 \cite{yolov2}    &      DarkNet-19        &      -           & 21.6   & 44.0         & 19.2          & 5.0           & 22.4           & 35.5        \\
   SSD512 \cite{ssd}       &      ResNet-101        &      -           & 31.2   & 50.4         & 33.3          & 10.2          & 34.5           & 49.8        \\
   RetinaNet \cite{retinanet} &   ResNet-101-FPN    &      -           & 39.1   & 59.1         & 42.3          & 21.8          & 42.7           & 50.2        \\
   Faster R-CNN \cite{fpn}  &     ResNet-101-FPN    &      -           & 36.2   & 59.1         & 42.3          & 21.8          & 42.7           & 50.2        \\
   Libra R-CNN \cite{libra}            &      ResNet-50-FPN    &      1x          & 38.7   & 59.9         & 42.0          & 22.5          & 41.1           & 48.7        \\
   Libra R-CNN \cite{libra}            &      ResNet-101-FPN    &     1x           & 40.3   & 61.3         & 43.9         & 22.9        & 43.1              & 51.0        \\
   Deformable R-FCN \cite{rfcn}&   Inception-ResNet-v2    &      -           & 37.5   & 58.0         & 40.8          & 19.4          & 40.1           & 52.5         \\
   Mask R-CNN \cite{mask-rcnn}&   ResNet-101-FPN    &      -           & 38.2   & 60.3         & 41.7          & 20.1          & 41.1           & 50.2         \\
   Grid-R-CNN \cite{gridrcnn}&   ResNet-101-FPN    &      2x           &  41.5  & 60.9         & 44.5          & 23.31          & 44.9          & 53.1         \\
   \hline
   RetinaNet*                   & ResNet-50-FPN       &      1x          & 35.9   &  55.9
          &     38.5          & 19.7            & 38.9              & 44.9            \\
   RetinaNet*                   & MobileNet-v2-FPN       &      1x          & 32.7   &  52.0  &  34.7       & 17.4            & 34.6     & 42.3            \\
   FCOS*                    &     ResNet-50-FPN       &      1x          & 37.0   &  56.6
          &  39.4             & 20.8            & 39.8              & 46.4            \\
   Faster R-CNN*           &      ResNet-50-FPN     &      1x          & 36.5   & 58.7         & 39.1          & 21.5          & 39.7           & 44.6         \\
   Faster R-CNN*           &      ResNet-101-FPN    &      1x          & 38.9   & 60.9         & 42.3          & 22.4          & 42.4           & 48.3         \\
   Faster R-CNN*           &      ResNet-101-FPN    &      2x          & 39.7   & 61.4         & 43.3          & 22.3          & 42.9           & 50.4          \\
   Faster R-CNN*           &      ResNext-101-32x4d-FPN&   1x          & 40.5   & 62.8         & 44.0          & 24.3          & 43.9           & 50.2          \\
   Faster R-CNN*           &      ResNext-101-64x4d-FPN&   1x          & 41.7   & 64.1         & 45.4          & 25.0          & 45.1           & 52.1          \\
   Faster R-CNN*           &      MobileNet-v2-FPN   &      1x          & 32.6      & 54.6            &34.3       & 18.6             & 34.5              & 41.0      \\
   Mask R-CNN*             &      ResNet-50-FPN     &       1x         & 37.5(34.4) & 59.4(56.3)& 40.6(36.6)   & 22.1(18.6)    & 40.6(37.2)     & 46.2(44.5)     \\
   Mask R-CNN*             &      ResNet-101-FPN     &      1x         & 39.8(36.3) & 61.6(58.5)& 43.3(38.7)   & 22.9(19.2)    & 43.2(39.3)     & 49.7(47.4)     \\
   Mask R-CNN*             &      ResNet-101-FPN     &      2x         & 40.8(36.9) & 62.2(59.1)& 44.6(39.6)   & 22.7(19.1)    & 44.0(39.9)     & 51.8(48.9)     \\
   \hline
   RetinaNet (ours)                  & ResNet-50-AugFPN       &      1x          & 37.5[\textbf{+1.6}]   &  58.4 &  40.1             & 21.3            & 40.5              & 47.3            \\
   RetinaNet (ours)                  & MobileNet-v2-AugFPN       &      1x          & 34.0[\textbf{+1.3}]   &  54.0    &  36.0             & 18.6            & 36.0              & 44.0    \\
   FCOS (ours)                   &     ResNet-50-AugFPN       &      1x          & 37.9[\textbf{+0.9}]   &  58.0 &  40.4             & 21.2            & 40.5              & 47.9            \\
   Faster R-CNN (ours)           &   ResNet-50-AugFPN     &      1x          & 38.8[\textbf{+2.3}]   & 61.5         & 42.0          & 23.3          & 42.1           & 47.7           \\
   Faster R-CNN (ours)           &   ResNet-101-AugFPN    &      1x          & 40.6[\textbf{+1.7}]   & 63.2         & 44.0          & 24.0          & 44.1           & 51.0             \\
   Faster R-CNN (ours)           &   ResNet-101-AugFPN    &      2x          & 41.5[\textbf{+1.8}]   & 63.9         & 45.1          & 23.8          & 44.7           & 52.8           \\
   Faster R-CNN (ours)           &   ResNext-101-32x4d-AugFPN&   1x          & 41.9[\textbf{+1.4}]   & 64.4         & 45.6          & 25.2          & 45.4           & 52.6           \\
   Faster R-CNN (ours)           &   ResNext-101-64x4d-AugFPN&   1x          & 43.0[\textbf{+1.3}]   & 65.6         & 46.9          & 26.2          & 46.5           & 53.9           \\
   Faster R-CNN (ours)           &    MobileNet-v2-AugFPN &      1x          & 34.2[\textbf{+1.6}]      & 56.6            & 36.2             & 19.6             & 36.4              & 43.1           \\
   Mask R-CNN (ours)           &   ResNet-50-AugFPN    &     1x           & 39.5[\textbf{+2.0}](36.3[\textbf{+1.9}]) & 61.8(58.7) & 42.9(38.8)  & 23.4(19.7)    & 42.7(39.2)     & 49.1(47.5)     \\
   Mask R-CNN (ours)          &   ResNet-101-AugFPN    &     1x           & 41.3[\textbf{+1.5}](37.8[\textbf{+1.5}]) & 63.5(60.4) & 44.9(40.4)  & 24.2(20.4)    & 44.8(41.0)     & 52.0(49.8)     \\
   Mask R-CNN (ours)          &   ResNet-101-AugFPN    &     2x          &  42.4[\textbf{+1.6}](38.6[\textbf{+1.7}]) & 64.4(61.4)  & 46.3(41.4)  & 24.6(20.6)    & 45.7(41.6)     & 54.0(51.4)     \\
   \hline
   \end{tabular}
   }
   \caption{Comparison with the state-of-the-art methods on COCO test-dev. The symbol '*' means our re-implementation results. For Mask R-CNN, the results in ( ) means the corresponding mask results. The number in [] stands for the  relative improvement. The training schedule follows the setting as Detectron \cite{detectron}.}
   \label{tab1}
   \end{table*}

\subsection{Residual Feature Augmentation}
   In FPN, feature map at the highest level $M_5$ is propagated in a top-down path and fused with the feature maps at lower levels $\left\{M_4, M_3, M_2\right\}$ gradually. On the one hand, feature maps of lower levels are enhanced with the semantic information from higher levels and the resulting features are endowed with diverse context information naturally. On the other hand, $M_5$ suffers from the information loss due to the reduced feature channels and only contains single scale context information that is not compatible with the resulting features at other levels.

   Based on this observation, we propose Residual Feature Augmentation to improve the feature representation of $M_5$ by utilizing a residual branch to instill diverse spatial context information into the original branch. We expect that the spatial context information can reduce the information loss in channels of $M_5$ and improves performance of the resulting feature pyramid simultaneously.
   To this end, we first produce multiple context features with different scales of ($\alpha_1 \times S, \alpha_2\times S,.., \alpha_n\times S$) by performing ratio-invariant adaptive pooling on $C_5$ whose scale is $S$. Then each context feature undergoes a $1\times1$ convolution layer independently to reduce feature channel dimension to 256. Finally, they are upsampled to a scale of $S$ via bilinear interpolation for subsequent fusion. Considering the aliasing effect caused by interpolation, we design a module named Adaptive Spatial Fusion (ASF) to adaptively combine these context features instead of simple summation. The detailed structure of ASF is illustrated in Figure \ref{fig2.1}. Specifically, ASF takes upsampled features as input and produces one spatial weight map for each feature. The weights are used to aggregate the context features into $M_6$, which is endowed with multi-scale context information.

   After $M_6$ is generated by ASF, it is combined with $M_5$ by summation and propagated to fuse with other features at lower levels. Finally, a $3\times3$ convolution layer is appended to each feature map to construct a feature pyramid $\left\{P_2, P_3, P_4, P_5\right\}$.

   Ratio-invariant adaptive pooling is different from PSP \cite{psp} in that PSP pools feature into multiple features with fixed sizes while ratio-invariant adaptive pooling takes the ratio of image into account, which is preferable to object detection. Furthermore, we fuse features with Adaptive Spatial Fusion instead of simple summation, which is inferior as shown in the experiments in ablation study.

\subsection{Soft RoI Selection}

   In FPN, feature for each RoI is obtained by pooling on one certain feature level, which is chosen according to the scale of that RoI heuristically. Generally, small RoIs are assigned to features of lower levels while large RoIs are assigned to that of higher levels. Under this strategy, two RoIs with similar sizes may be assigned to different levels. This can produce sub-optimal results because it is ambiguous which feature level contains the most important information of an RoI. It is challenging to design a perfect strategy to allocate the RoIs.

   PANet \cite{panet} addresses this by pooling RoI features from all levels and using the maximum of RoI features adapted by fully connected layers to refine the proposals. It improves the performance of instance segmentation but the extra fully connected layers increase the parameters significantly. Furthermore, the max operation only selects the feature points with the highest responses and ignores the features with lower responses in other levels that may be also beneficial for recognition. This may impedes the features at different levels from being fully exploited. Therefore, we propose Soft RoI Selection, which learns to generate better RoI features from features at all pyramid levels by parameterizing the procedure of RoI pooling. Soft RoI Selection introduces adaptive weights to better measure the importance of feature inside the RoI region at different levels. The final RoI features are generated based on the adaptive weightes instead of the hard selection approaches like RoI assignment or max operation.

   Specifically, we first pool features from all pyramid levels for each RoI. Then instead of adapting the RoI features with fully connected layers like PANet, we exploit an Adaptive Spatial Fusion module (ASF), which is also a component in Residual Feature Augmentation, to fuse these features adaptively. It generates different spatial weight maps for RoI features from different levels and the RoI features are fused with weighted aggregation. ASF only consists of two convolution layers and consumes much fewer parameters than the extra fully connected layers used in PANet. In this way, Soft RoI Selection parameterizes the procedure of RoI pooling. It can be leaned by back-propagation with other components in the network and does not rely on a heuristically designed strategy.

\section{Experiments}
\subsection{Dataset and Evaluation Metrics}
 We perform all experiments on the MS COCO detection dataset with 80 categories. It contains 115k images for training (\textit{train}2017), 5k images for validation (\textit{val}2017) and 20k images for testing (\textit{testdev}). The labels of \textit{testdev} are not released publicly. We train models on \textit{train}2017 and report results of ablation study on \textit{val}2017. The final results are reported on \textit{testdev}. All reported results follow standard COCO-style Average Precision (AP) metrics.

\subsection{Implementation Details}

 All experiments are implemented based on mmdetecton \cite{mmdet}. The input images are resized to have a shorter size of 800 pixels. By default, we train the models with 8 GPUs (2 images per GPU) for 12 epochs. The initial learning rate is set as 0.02 and it decreases by a ratio of 0.1 after the 8th and 11th epoch respectively. $\lambda$ in Equ. \ref{equ1} is set as 0.25; as for the setting of ratio-invariant adaptive pooling, three alphas $\alpha_1, \alpha_2, \alpha_3$ with values as 0.1, 0.2, and 0.3 respectively are chosen if not noted specifically. All other hyper-parameters in this paper follow mmdetection.

\subsection{Main Results}

 In this section, we evaluate AugFPN on COCO \textit{testdev} set and compare with other state-of-the-art one-stage and two-stage detectors. For a fair comparison, we re-implement the corresponding baseline methods equipped with FPN. All results are shown in Table \ref{tab1}. By replacing FPN with AugFPN, Faster R-CNN using ResNet50 as backbone (denoted as ResNet50-AugFPN) achieves 38.8 AP, which is 2.3 points higher than Faster R-CNN based on ResNet50-FPN. Besides, AugFPN can consistently bring non-negligible performance even with more powerful backbone networks. For example, when using ResNext101-32x4d and ResNext101-64x4d as the feature extractors, our method still improves the performance by 1.4 and 1.3 AP, respectively.

 Obviously, Faster R-CNN with AugFPN significantly improves FPN when using powerful models like ResNet50 as backbone. Now we test whether AugFPN is suitable for light-weight models, \textit{i.e.} MobileNet-V2 \cite{mobilenetv2}. As shown in Table \ref{tab1}, Faster R-CNN with MobileNet-v2-AugFPN outperforms MobileNet-v2-FPN by 1.6 AP under 1$\times$ learning rate schedule.

 As for one-stage detectors, we validate the effectiveness of AugFPN on two different types of detectors, \textit{i.e.} anchor-based RetinaNet \cite{retinanet} and anchor-free FCOS \cite{fcos}. Since no concept of RoIs exist in these two detectors, Soft RoI Selection is not included in this case. Therefore, the outputs of detectors instead of RPN are used by the Consistent Supervision module in the training phase. As shown in Table \ref{tab1}, RetinaNet can be improved by 1.6 AP and 1.3 AP respectively when using ResNet50 or MobileNet-v2 as backbone. Meanwhile, FCOS is boosted to 37.9 AP from 37.0 AP when replacing FPN with AugFPN. The improvements show that the other two components still improve the feature representation of feature pyramid a lot even without including Soft RoI Selection.

 Finally, we evaluate AugFPN on Mask R-CNN. By replacing FPN with AugFPN, Mask RCNN with ResNet50 is improved by 2.0 AP on the detection and 1.9 AP on instance segmentation. When using ResNet101 as backbone, the improvement of AugFPN reaches 1.5 AP on the detection and 1.5 AP on instance segmentation respectively. As can be seen in Table \ref{tab1}, AugFPN brings consistent improvements on various backbones, detectors and even different tasks. This verifies the robustness and generalization ability of AugFPN.

\begin{table}
    \centering
    \resizebox{1.0\columnwidth}{!}{
    \begin{tabular}{ccccccccc}
    \hline
    \textbf{CS}     &    \textbf{RFA}     &  \textbf{SRS}       &   AP    &   $AP_{50}$     &    $AP_{75}$        &    $AP_{s}$      &    $AP_{m}$      &    $AP_{l}$\\
    \hline
      &  &  &  36.3  & 58.3  &  39.0   & 21.4  & 40.3 & 46.6  \\
    \hline
    \cmark &  &  & 37.2 & 59.2 & 40.1 & 21.8 & 40.9 & 47.8  \\
     & \cmark & & 37.3 & 59.8 & 40.4 & 22.5  & 41.3 & 47.2\\
     &  & \cmark & 37.1 & 59.1 & 40.1 & 21.8  & 41.3 & 47.5\\
     \hline
    \cmark & \cmark &  & 37.7 & 60 & 40.8 & 22.8 & 41.4 & 48.4  \\
    \cmark  &   & \cmark  & 38.0 & 60.3 & 41.5 & 22.9 & 41.9 & 48.0  \\
     & \cmark & \cmark & 37.9 & 60.3 & 40.7 & 23.6 & 41.8 & 47.9  \\
    \hline
    \cmark  & \cmark  & \cmark & 38.7 & 61.2 & 41.9 & 24.1 & 42.5 & 49.5  \\
    \hline
    \end{tabular}
    }
    \caption{Effect of each component. Results are reported on COCO \textit{val}2017. \textbf{CS}: Consistent Supervision, \textbf{RFA}: Residual Feature Augmentation, \textbf{SRS}: Soft RoI Selection}
    \label{tab2}
\end{table}

\subsection{Ablation Study} \label{ablation}

In this section, we conduct extensive ablation experiments to analyze the effects of individual components in our proposed method.
\vspace{-0.2cm}
\paragraph{Ablation studies on importance of each components.} To analyze the importance of each component in AugFpn, Consistent Supervision, Residual Feature Augmentation and Soft RoI Selection are gradually applied to the model to validate the effectiveness. Meanwhile, the improvements brought by combination of different components are also presented to demonstrate that these components are complementary to each other. The baseline method for all ablation studies is Faster R-CNN with ResNet50-FPN. All results are shown in Table \ref{tab2}.

As shown in Table \ref{tab2}, Consistent Supervision improves the baseline method by 0.9 AP. This benefits from that Consistent Supervision narrow semantic gaps between the features after lateral connection and improves their semantic representation simultaneously. It is worthy to note that Consistent Supervision does not introduce extra parameters in inference. Therefore it is cheap to add it to any other FPN based detection models.

Residual Feature Augmentation improves the detection performance from 36.3 to 37.3 AP. It can be seen that results of objects in small, medium and large scale are all improved, which means the complementary information added to $M_5$ also benefits the feature maps at lower levels and improves the feature representation of feature pyramid simultaneously.

Soft RoI Selection brings 0.8 AP improvement to the baseline method. Specifically, the improvements of $AP_m$ (+1.0 AP) and $AP_l$ (+0.9 AP) contribute most to the final improvement. These results indicate that adaptive spatial fusion enables larger RoIs, which are originally assigned to higher feature levels, to incorporate features from lower levels that contain more information of spatial details.
 \begin{table}
    \centering
    \resizebox{1\columnwidth}{!}{
    \begin{tabular}{ccccccccc}
    \hline
    Setting      &    $\lambda$     &    AP    &   $AP_{50}$     &    $AP_{75}$        &    $AP_{s}$      &    $AP_{m}$      &    $AP_{l}$\\
    \hline
    no supervision &   0.0          & 36.3  & 58.3  &  39.0   & 21.4  & 40.3 & 46.6  \\
    \hline
    single level   &   1.2        &  36.7    &  58.5    &  39.7      &  21.3    &  40.1   &  47.3     \\
    single level   &   1.0        & 37.0  & 58.9  & 40.2    & 21.8  & 40.4 & 47.5   \\
    single level   &   0.5        & 36.9  & 58.7  & 40.0    & 21.7  & 40.9 & 47.4   \\
    single level   &   0.25       & 36.7  & 58.7  & 39.8    & 21.5  & 40.3 & 47.2   \\
    \hline
    all level      &   0.5        & 36.9  & 58.8  & 39.9    & 21.8  & 40.7 & 47.1   \\
    all level      &   0.25       & 37.2  & 59.2  & 40.1    & 21.8  & 40.9 & 47.8   \\
    all level      &   0.125      & 37.1  & 58.9  & 40.1    & 22.3  & 40.9 & 47.4   \\
    \hline
    \end{tabular}
    }
    \caption{Ablation studies of Consistent Supervision on COCO \textit{val}2017.}
    \label{tab3}
\end{table}

When combining any two of three components, the improvement over the baseline method is much higher. For example, Consistent Supervision and Soft RoI Selection together can lead to 1.7 AP improvement. When three components are all integrated into the baseline method, it can achieve 38.7 AP with 2.4 AP improvement. These results indicate that these three components are complementary to each other and tackle different problems in FPN.

\begin{table}
    \centering
    \resizebox{1.0\columnwidth}{!}{
    \begin{tabular}{ccccccccccc}
    \hline
    Setting      &    Pooling Type    & $\alpha$  &     AP    &   $AP_{50}$     &    $AP_{75}$        &    $AP_{s}$     &    $AP_{m}$       &    $AP_{l}$\\
    \hline
    baseline      &   -         &   -      &    36.3  & 58.3  &  39.0   & 21.4  & 40.3 & 46.6  \\
    \hline
    sum           &      GMP          &    -     &     34.5  & 56.6  & 36.8    & 21.9  & 38.3 & 42.4   \\
    sum           &      GAP          &    -      &     36.8  & 59.3  & 39.7    & 22.1  & 40.9 & 46.7   \\
    sum           &   RA-AP           &  0.1,0.2,0.3 &  37.1  & 59.8  & 39.9    & 22.7  & 41.1 & 47.3   \\
    \hline
    ASF           &    RA-AP          & 0.1     &     37.1   &   59.6       &  40.2    &  22.3     & 40.9   &  47.2     \\
    ASF           &    RA-AP          & 0.1,0.2   &   37.2   &  59.4    &  40.1     & 22.4  &  41.1   & 47.7   \\
    ASF           &    RA-AP          &  0.1,0.2,0.3 &  37.3     & 59.8  & 40.4    & 22.5  & 41.1 & 47.4   \\
    ASF           &    RA-AP          & 0.1,0.2,0.3,0.4   &   37.4       &  59.9    &  40.5     & 22.5     &  41.1    &   47.9 \\
    \hline
    ASF           &    RA-AP          &  0.1,0.2,0.4 &  37.3  & 59.7  & 40.2    & 22.9  & 41.3 & 47.2   \\
    ASF           &    RA-AP          &  0.1,0.2,0.5 &  37.2  & 59.7  & 40.3    & 22.2  & 41.1 & 47.0   \\
    ASF           &    PSP          &  -              &  37.0   & 59.5  & 40.1    & 22.8  & 40.9 & 47.3   \\
    \hline
    \end{tabular}
    }
    \caption{Ablation studies of Residual Feature Augmentation on COCO \textit{val}2017. GMP, GAP means Global Max Pooling and Global Average Pooling respectively. RA-AP means ratio-invariant average pooling. ASF means Adaptive Spatial Fusion.}
    \label{tab4}
\end{table}

\paragraph{Ablation studies on Consistent Supervision.} Experiment results related with three settings of Consistent Supervision are presented in Table \ref{tab3}. The first setting is the baseline method, where $\lambda$ in Equ. \ref{equ1} is set as zero. The second setting is \textit{single level} supervision, which only applies supervision signals to the feature map that RoIs are assigned to according to the assignment strategy of RoIs in FPN \cite{fpn}. The third setting is \textit{all level} supervision, which enforces supervision signals to feature maps of all levels.

When using \textit{single level} supervision, the baseline method can be improved by 0.7 AP by setting $\lambda$ as 1.0. The improvement becomes smaller when $\lambda$ is set as other values. By applying supervision signals on feature maps at all levels, \textit{all level} supervision obtains better results than both \textit{single level} setting and baseline method. It can be seen that when $\lambda$ is set as 0.25, \textit{all level} setting brings 0.5 and 0.9 AP improvement than \textit{single level} setting and baseline model, respectively. The superiority of \textit{all level} setting verifies that forcing feature maps at all levels to learn similar semantic information is an effective practice to narrow the semantic gap between them and improves the performance of the resulting feature pyramid.

\paragraph{Ablation studies on Residual Feature Augmentation.}
The results of ablation studies on Residual Feature Augmentation are shown in Table \ref{tab4}. We first explore the influence of pooling type by using global pooing instead of ratio-invariant adaptive pooling. Since there is only one branch, Adaptive Spatial Fusion (ASF) is not adopted. Two types of global pooling, Global Max Pooling (GMP) and Global Average Pooling (GAP), are tested in the experiments. From the results shown in Table \ref{tab4}, we observe that GMP is inferior to GAP. GAP improves the baseline method by 0.6 AP while GMP degrades the accuracy instead, which means average pooling is more robust than max pooling because output of max pooling may be disturbed by the peak noises in feature maps greatly.

\begin{figure*}
    \centering
    \includegraphics[width=0.9\textwidth]{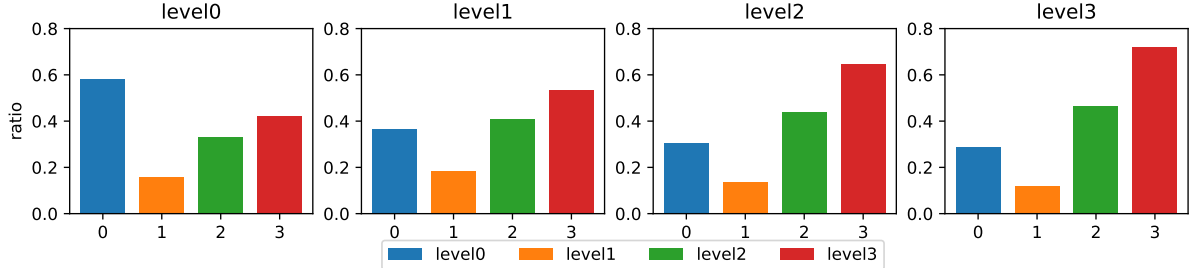}
    \caption{Ratios of features pooled from each pyramid level with Soft RoI Selection. The figures from left to right correspond to the RoIs originally assigned to $P_2-P_5$. The results are obtained on COCO \textit{val}2017.}
    \label{fig3}
\end{figure*}

Based on this observation, we replace GAP with ratio-invariant adaptive average pooling (RA-AP). We firstly choose an $\alpha$ setting of three alphas with values as 0.1, 0.2 and 0.3 respectively. The influence of different $\alpha$ setting will be discussed afterward. For a fair comparison, the pooled context features are directly fused by summation instead of ASF. As shown in the fourth row in Table \ref{tab4}, RA-AP brings 0.8 AP and 0.3 AP improvement over the baseline method and GAP, which validates the effectiveness of diverse context brought by the residual branch.
By combining ASF with RA-AP using the same $\alpha$ setting, the final result can be further boosted to 37.3 AP, which is 1.0 AP higher than the baseline method.

The influence of different $\alpha$ setting is also investigated. Although mAP increase as the number of $\alpha$ increases, as can be seen from Table \ref{tab4}, our final model adopts the setting of three alphas for a better trade-off between complexity and accuracy. In addition, we explore how different $\alpha$ values impact the performance and the experimental results are shown in the fourth part of table \ref{tab4}. When values of $\alpha$ are set as other values, the performance is even worse or shows no more improvement. To further validate the effectiveness of RA-AP, we replace RA-AP with PSP \cite{psp} that pools feature map into fixed sizes $1\times1, 2\times2, 3\times3$. The experimental result shows that it is inferior to RA-AP by 0.3 AP, which verifies that ratio-invariant adaptive pooling can preserve more information beneficial for recognition by not disturbing the original ratio of the objects in features.

\begin{table}
    \centering
    \resizebox{1.0\columnwidth}{!}{
    \begin{tabular}{ccccccccc}
    \hline
    Setting    &   Fusion type &    AP    &   $AP_{50}$     &    $AP_{75}$        &    $AP_{s}$      &    $AP_{m}$      &    $AP_{l}$\\
    \hline
    \multicolumn{2}{c}{baseline}           & 36.3  & 58.3  &  39.0   & 21.4  & 40.3 & 46.6  \\
    \hline
    SRS   & sum      & 36.6  & 59.0  & 39.1    & 22.3  & 40.6 & 46.4   \\
    SRS    & max     & 36.5  & 58.5  & 39.2    & 21.6  & 40.2 & 46.9   \\
    SRS    & ACF   & 37.0  & 59.2  & 39.8    & 22.0  & 41.2 & 46.8   \\
    SRS    & ASF   & 37.1  & 59.1  & 40.1    & 21.8  & 41.3 & 47.5   \\
    \hline
    \end{tabular}
    }
    \caption{Ablation studies of Soft RoIs Selection on COCO \textit{val}2017. SRS, ACF and ASF is the acronym of Soft RoI Selection, Adaptive Channel Fusion and Adaptive Spatial Fusion respectively. }
    \label{tab5}
\end{table}

\paragraph{Ablation studies on Soft RoI Selection.}
We first study different methods of fusing RoI features. The first one is sum fusion and the second one is max fusion. The only difference between max fusion in this setting and adaptive pooling in PANet \cite{panet} is that we do not introduce extra fully connected layers to adapt the RoI features because it would significantly increases the parameters. The third one is the Adaptive Channel Fusion (ACF) as shown in Figure \ref{fig2.2}. It is inspired by the SE module \cite{senet} but with a different goal of fusing different RoI features from the perspective of channel importance. The fourth one is the Adaptive Spatial Fusion (ASF) module as shown in Figure \ref{fig2.1}. Experimental results on these methods are shown in Table \ref{tab5}.

From the results we can observe that sum fusion and max fusion improve baseline method by 0.3 and 0.2 AP respectively. By using ACF to fuse RoI features adaptively, the baseline method obtains 0.7 AP improvement. When ACF is replaced with ASF, which is the setting of Soft RoI Selection, the final model achieves 37.1 AP and outperforms the baseline method by 0.8 AP. These results indicate that by enabling the procedure of RoI feature selection to learn with other components, Soft RoI Selection can produce more powerful representations of RoIs.

In order to analyze the ratios of features at different levels absorbed by ASF, we divide RoI proposals on \textit{val}2017 into four levels according to the levels they are originally assigned to. For each RoI, we average over all positions on each weight map generated by ASF and obtain four ratios corresponding to four feature levels. Finally, for all RoIs that belong to one certain level, four ratio values are separately averaged over these RoIs. The results corresponding to four pyramid levels are illustrated in Figure \ref{fig3}. Obviously, features from all levels contribute together to generate better RoI features, which indicates that features from all levels are beneficial for the recognition of each RoI. It can be seen that the RoIs originally assigned to level P2 still requires more semantic information from P5 beside the information propagated from higher levels. Meanwhile, the RoIs originally assigned to P3-5 all requires much detailed appearance information from P2, which may be lost due to down-sampling.

\subsection{Runtime Analysis}
We also measure the time of training and testing when FPN is replaced with AugFPN. Specifically, the training time of Faster-RCNN with ResNet50-AugFPN is about 1.1 hour and that of Faster-RCNN with ResNet50-FPN is nearly 0.9 hour for each epoch on COCO dataset with the same batch size of 16. As for the inference time, AugFPN can run at 11.1 fps and FPN can run at 13.4 fps for images with a shorter size of 800 pixels. The inference time is the average inference time over COCO \textit{val5000} split including the time of data loading, network forwarding, and post-processing. All the runtimes are tested on Tesla V100.

\section{Conclusion}
In this paper, we analyze the inherent problems along with FPN and find that the multi-scale features are not fully exploited. Based on this observation, we propose a new feature pyramid network named AugFPN to further exploit the potential of multi-scale features. By integrating three simple yet effective components, \textit{i.e.} Consistent Supervision, Residual Feature Augmentation, and Soft RoI Selection, AugFPN can improve the baseline method by a large margin on the challenging MS COCO dataset.

{\small
\bibliographystyle{ieee_fullname}
\bibliography{egbib}
}

\end{document}